\documentclass[preprint]{article}

\usepackage[sort&compress,numbers]{natbib}

\usepackage{neurips_2025}

\usepackage[utf8]{inputenc} 
\usepackage[T1]{fontenc}    
\usepackage{hyperref}
\usepackage{url}            
\usepackage{booktabs}       
\usepackage{amsfonts}       
\usepackage{nicefrac}       
\usepackage{microtype}      

\usepackage{graphicx}
\usepackage{wrapfig}
\usepackage{caption}
\usepackage{subcaption}
\usepackage{amsmath,amsthm}
\newtheorem{theorem}{Theorem}
\newtheorem{corollary}[theorem]{Corollary}
\newtheorem{proposition}[theorem]{Proposition}
\newtheorem{lemma}[theorem]{Lemma}
\newtheorem{definition}[theorem]{Definition}

\title{GrokAlign: Geometric Characterisation and Acceleration of Grokking} 

\author{%
  Thomas Walker$^1$\thanks{Correspondence to \texttt{tw78@rice.edu}}
  \And
  Ahmed Imtiaz Humayun$^2$\thanks{Work done before joining Google.}
  \And
  Randall Balestriero$^3$
  \And
  \vspace{0.5em}
  Richard Baraniuk$^1$\\
  $^1$Department of Electrical and Computer Engineering, Rice University\\$^2$Google Research\\
  $^3$Department of Computer Science, Brown University
}

\begin{document}

\maketitle

\begin{abstract}
    A key challenge for the machine learning community is to understand and accelerate the training dynamics of deep networks that lead to delayed generalisation and emergent robustness to input perturbations, also known as grokking. Prior work has associated phenomena like delayed generalisation with the transition of a deep network from a linear to a feature learning regime, and emergent robustness with changes to the network's functional geometry, in particular the arrangement of the so-called linear regions in deep networks employing continuous piecewise affine nonlinearities. Here, we explain how grokking is realised in the Jacobian of a deep network and demonstrate that aligning a network's Jacobians with the training data (in the sense of cosine similarity) ensures grokking under a low-rank Jacobian assumption. Our results provide a strong theoretical motivation for the use of Jacobian regularisation in optimizing deep networks -- a method we introduce as GrokAlign -- which we show empirically to induce grokking much sooner than more conventional regularizers like weight decay. Moreover, we introduce centroid alignment as a tractable and interpretable simplification of Jacobian alignment that effectively identifies and tracks the stages of deep network training dynamics. Accompanying \href{https://thomaswalker1.github.io/blog/grokalign.html}{webpage} and \href{https://github.com/ThomasWalker1/grokalign}{code}.
\end{abstract}

\begin{figure}[h]
  \centering
  \begin{subfigure}{0.3\columnwidth}
    \includegraphics[width=0.9\textwidth]{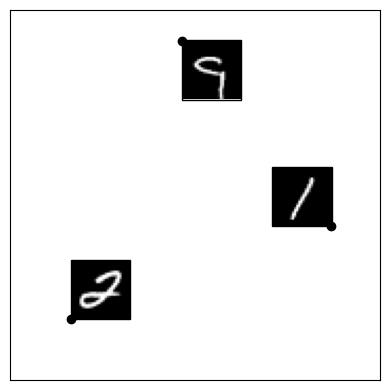}
    \centering
     \caption*{Data Points}
 \end{subfigure}
  \hspace{0.5em}
  \begin{subfigure}{0.3\columnwidth}
    \includegraphics[width=0.9\textwidth]{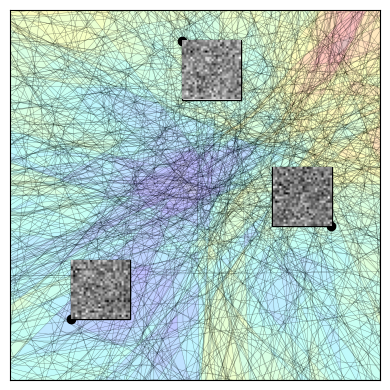}
    \centering
     \caption*{Centroids at Memorisation}
 \end{subfigure}
  \hspace{0.5em}
  \begin{subfigure}{0.3\columnwidth}
    \includegraphics[width=0.9\textwidth]{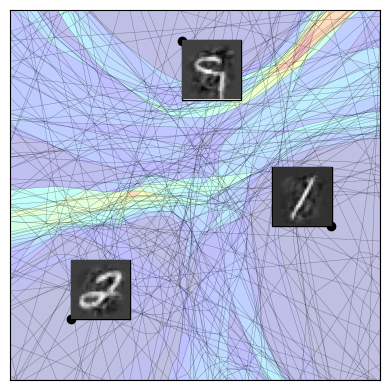}
    \centering
     \caption*{Centroids at Generalisation}
 \end{subfigure}
  \caption{\small For a deep network to grok, its Jacobians should \emph{align} such that the sum of their rows are cosine-similar to the point at which they were computed; we dub this condition {\em centroid aligned}. We train a ReLU network on the MNIST dataset \citep{lecun_gradient-based_1998} using GrokAlign. We take three training data points, \textbf{left}, and observe the linear regions (using SplineCam \citep{humayun_splinecam_2023}) of the deep network along with the centroids \citep{balestriero_geometry_2019} of the three data points when it has memorized the training data, \textbf{centre}, and when it has generalised, \textbf{right}. We colour the linear regions according to the norm of the linear operator acting upon them. (Additional figures can be found in Figure \ref{fig:centroid_alignment2}).}
  \label{fig:centroid_alignment}
\end{figure}

\section{Introduction}

Deep networks are known to have emergent properties during prolonged training that are essential to understand to facilitate their reliable and effective training. 
\emph{Delayed generalisation} involves the test accuracy increasing long after train accuracy has saturated, a phenomenon initially termed \emph{grokking} \citep{power_grokking_2022}. 
Delayed generalisation spans multiple deep architectures and domains, including transformers on algorithmic tasks \citep{power_grokking_2022} and natural language processing \citep{wang_loss_2024}, and fully connected networks performing image-classification \citep{liu_omnigrok_2022}. 
Subsequently, the grokking concept has been expanded to include \emph{delayed robustness} \citep{humayun_deep_2024}, which involves prolonged training inducing a robustification of the deep network to input perturbations. 
Ideally we would accelerate the onset of both generalisation and robustness in deep networks, though in practice there has been an observed tension between them \citep{tsipras_robustness_2019,zhang_theoretically_2019}.

Despite a high-level understanding of grokking, there exists no foundational explanation for why it occurs nor a practical and interpretable framework for accelerating a deep network's training dynamics to reach the grokked state more efficiently.

Prior work in this space attributes delayed generalisation to the transition of a deep network from a linear to a feature learning regime \citep{lyu_dichotomy_2024,rubin_grokking_2024,kumar_grokking_2024}. 
Studies have explored this dynamic from the perspective of the neural tangent kernel \citep{jacot_neural_2018,kumar_grokking_2024}, the adaptive kernel \citep{seroussi_separation_2023,rubin_grokking_2024}, and mechanistic interpretability \citep{nanda_progress_2022,varma_explaining_2023}. On the one hand, these works have arrived at an array of sufficient conditions for inducing generalisation, including weight-norm at initialisation \citep{liu_omnigrok_2022}, weight-decay \citep{nanda_progress_2022,lyu_dichotomy_2024}, dataset size \citep{varma_explaining_2023}, and output or label scaling \citep{kumar_grokking_2024}. 
On the other hand, delayed generalisation and robustness has been attributed to the evolution of the functional geometry of the network \citep{humayun_deep_2024}, which is a reference to the arrangement of the so-called linear regions of a continuous piecewise affine network. 
The Jacobian matrices of a deep network have also been identified as intimately related to their robustness \citep{rifai_contractive_2011,jakubovitz_improving_2018,hoffman_robust_2019,etmann_connection_2019,chan_jacobian_2020}.

{\em In this paper, we prove theoretically and demonstrate empirically that Jacobian norm constraints induces grokking in deep networks.
Moreover, we develop a tractable and interpretable approach for monitoring and accelerating grokking in practice based on an efficient summarization of the Jacobian called the centroid.}

This paper makes three main contributions.
First, we demonstrate that deep networks that have optimized their loss function have \emph{aligned} Jacobians at the training data points, in the sense that the rows of the Jacobians at the training points are simply scalar multiples of those points.
Deep networks with aligned Jacobians have been empirically demonstrated to be robust \citep{etmann_connection_2019,chan_jacobian_2020}, and we prove rigorously that they are optimally robust amongst all rank-one Jacobians. 
Since deep network training dynamics tend to bias the Jacobian towards a low-rank matrix \citep{le_training_2022,huh2023simplicitybias,timor_implicit_2023,yunis2024rank,galanti_sgd_2025}, we conclude that the \emph{the cause of grokking is the alignment of the deep network's Jacobian matrices}. 
Through this theory we introduce \emph{GrokAlign}, which is the use of Jacobian regularisation to ensure and accelerate grokking.

Second, since working with Jacobian matrices in practice is computationally expensive, and current strategies to align the Jacobians are cumbersome \citep{chan_jacobian_2020}, we propose to summarize the Jacobian matrices via the sum of their rows.
This vector can be efficiently computed through Jacobian vector products \citep{balestriero_fast_2021}, and it has a strong geometrical interpretation in the spline theory of deep learning \citep{balestriero_spline_2018}, where it corresponds to the \emph{centroid} of the linear region containing the data point of interest. Theoretically, the centroids are connected to the neural tangent kernel, and empirically they offer efficiently computable metrics for monitoring and detecting the emergence of grokking.

Third, we explore the practical significance of Jacobian and centroid alignment as a framework through which to consider the dynamics of deep network training. We demonstrate that centroid alignment can reliably identify key phases in the training dynamics of deep networks and when additional training could be beneficial. Furthermore, we illustrate the effectiveness of GrokAlign as a strategy for controlling the training dynamics of deep networks that lead to grokking.

In summary, in Section \ref{sec:jacobian_regularisation_ensures_grokking}, we theoretically identify Jacobian alignment as the grokked state of a deep network which can be arrived at by regularising the Jacobian matrices of the deep network during training -- a method we introduce as GrokAlign. In Section \ref{sec:dynmamics_of_deep_network_centroids}, we demonstrate how we can equivalently track Jacobian alignment through centroid alignment. In Section \ref{sec:experiments}, we put this into practice by demonstrating that we can use centroid alignment to identify the key phases of deep network training and we can use GrokAlign to induce robustness, inhibit or accelerate grokking and control the learned solutions of deep networks.

\section{Jacobian Regularisation Explains Grokking}\label{sec:jacobian_regularisation_ensures_grokking}

Let $f:\mathbb{R}^d\to\mathbb{R}^C$ be a deep network. 
Here we focus on the classification setting so that the prediction of the deep network at a point $\mathbf{x}$ is taken to be $\mathrm{argmax}(f(\mathbf{x}))$. In this setting, the deep network is trained on a data set $\left\{\left(\mathbf{x}_p,y_p\right)\right\}_{p=1}^m$ -- where $\mathbf{x}_p\in\mathbb{R}^d$ and $y_p\in\mathbb{R}$ is its corresponding class -- under some loss function $\mathcal{L}=\frac{1}{m}\sum_{p=1}^m\ell\left(f\left(\mathbf{x}_p\right),y_p\right)$. Let $J_{\mathbf{x}}(f)$ be the Jacobian of $f$ at $\mathbf{x}$.

\begin{definition}\label{def:jacobian_aligned}
    A deep network is \textbf{Jacobian-aligned} at $\mathbf{x}\in\mathbb{R}^d$ if $J_{\mathbf{x}}(f)=\mathbf{c}\mathbf{x}^\top$ for some vector $\mathbf{c}\in\mathbb{R}^d$.
\end{definition}

\begin{wrapfigure}{r}{0.4\textwidth}
  \begin{center}
    \includegraphics[width=0.4\textwidth]{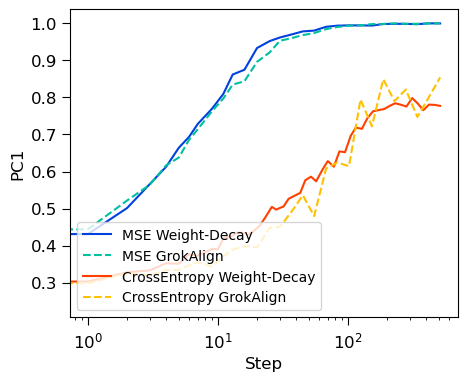}
  \end{center}
  \caption{\small Under weight-decay and GrokAlign, the effective rank of the Jacobian matrices evaluated at the training data tends towards rank one. Here we trained ReLU networks on the MNIST classification task \citep{lecun_gradient-based_1998} under the mean-squared error and cross-entropy loss functions using the AdamW optimizer \citep{loshchilov_decoupled_2019}. Throughout training, we recorded the average explained variance of the first principal component of the Jacobians evaluated at the training data (PC1), namely $\frac{\sigma_1^2}{\sum_{i=1}^r\sigma_i^2}$ where $\sigma$ are the singular values of the Jacobian. When this normalized value equals one, the Jacobian matrix is rank one.}
  \label{fig:pc1}
  \vspace{-2.5em}
\end{wrapfigure}

A deep network is said to have \emph{generalised} when it learns to extrapolate beyond the training set and perform well on unseen inputs, and it is said to be \emph{robust} when applying perturbations to inputs does not change the behaviour of the deep network drastically. The emergence of generalisation is typically formalised under the feature learning regime of training \citep{lyu_dichotomy_2024,rubin_grokking_2024,kumar_grokking_2024}, whilst the robustness of deep networks has been connected to its Jacobians \citep{etmann_connection_2019,chan_jacobian_2020}. The delayed onset of both these properties is encapsulated in the grokking phenomenon \citep{power_grokking_2022,humayun_deep_2024}.

Let us suppose that $f$ is a continuous piecewise affine deep network -- which includes a broad class of architectures, including ReLU feedforward networks, recurrent networks, convolutional neural networks, and residual networks with piecewise linear activation functions \citep{balestriero_spline_2018}. Such a deep network has a representation of the form $f(\mathbf{x})=A_{\omega_{\mathbf{x}}}\mathbf{x}+B_{\omega_{\mathbf{x}}}$ where $A_{\omega_{\mathbf{x}}}\in\mathbb{R}^{C\times d}$ and $B_{\omega_{\mathbf{x}}}\in\mathbb{R}^C$. More specifically, $A_{\omega_{\mathbf{x}}}$ and $B_{\omega_{\mathbf{x}}}$ are the parameters for the affine transformation operating on the linear region $\omega_{\mathbf{x}}$ encompassing $\mathbf{x}$. The {\em functional geometry} of $f$ is the disjoint union of these linear regions, which is a finite-partition of the input space into a collection of convex polytopes \citep{balestriero_geometry_2019}. Note that in this setting $J_{\mathbf{x}}(f)=A_{\omega_{\mathbf{x}}}$.

\begin{theorem}\label{thm:optimal_jacobian}
    Let $\mathcal{L}$ be the cross-entropy or mean-squared error loss function. Then the continuous piecewise affine deep network $f$ minimising $\mathcal{L}$ under the constraints that $\left\Vert J_{\mathbf{x}_p}(f)\right\Vert_F^2\leq\alpha$ and $B_{\omega_{\mathbf{x}_p}}=\mathbf{0}$ for every $p=1,\dots,m$, is Jacobian-aligned. (Proof in Appendix \ref{proof:optimal_jacobian}).
\end{theorem}

Theorem \ref{thm:optimal_jacobian} demonstrates that Jacobian-aligned deep networks are optimal in the sense of optimising the training objective. Combined with prior works \citep{etmann_connection_2019,chan_jacobian_2020}, we can also infer that Jacobian-aligned deep networks are robust.\footnote{Although, in these prior works, the notion of alignment considers only the row of the Jacobian matrix corresponding to the class of the input.} We support this with the following.

\begin{theorem}\label{thm:optimally_robust}
    If $A_{\omega_{\mathbf{x}}}$ is a rank-one matrix and $B_{\omega_{\mathbf{x}}}=\mathbf{0}$, then the local mapping on the linear region $\omega_{\mathbf{x}}$ is optimally robust with respect to $\ell_2$ perturbations when $A_{\omega_{\mathbf{x}}}=\mathbf{c}\mathbf{x}^\top$, where the maximum entry of $\mathbf{c}$ is at the index of the class of $\mathbf{x}$. (Proof in Appendix \ref{proof:optimally_robust}).
\end{theorem}

In practice, the dynamics of deep network training biases toward low rank weight matrices \citep{le_training_2022,huh2023simplicitybias,timor_implicit_2023,yunis2024rank,galanti_sgd_2025}, and thus low rank Jacobians (see Figure \ref{fig:pc1}). Hence, from Theorem \ref{thm:optimally_robust}, we determine that delayed robustness ought to necessarily involve the Jacobian alignment of deep networks.

By explicitly enforcing the Jacobian norm constraint of Theorem \ref{thm:optimal_jacobian} -- a method we introduce as GrokAlign -- we can guarantee that optimising the training objective will lead to a grokked network. More specifically, GrokAlign involves appending the average Frobenius norm\footnote{The Frobenius norm of a matrix is equal to the square root of the sum of the squares of its components.} of the Jacobian matrices at the training data to the loss function with some weighting coefficient, $\lambda_{\text{Jac}}$.

Reassuringly, this form of Jacobian regularisation has been demonstrated to improve the robustness of deep networks in prior work \citep{rifai_contractive_2011,gu_towards_2014,jakubovitz_improving_2018,hoffman_robust_2019}. Moreover, forcing the network to maintain a low Lipschitz constant, which is equivalent to having a low Jacobian norm, has been identified to balance the observed trade-off between generalisation and robustness \citep{yang_closer_2020}.

Weight-decay could also be a viable strategy of enforcing the constraint of Theorem \ref{thm:optimal_jacobian}; however, it is less direct. Indeed, in some cases weight-decay has proven effective for inducing grokking \citep{power_grokking_2022,nanda_progress_2022,varma_explaining_2023,lee_grokfast_2024}, while in other cases it has shown to be insufficient for grokking \citep{kumar_grokking_2024}.

In summary, we have identified that grokking requires the alignment of a deep network's Jacobian which, due to Theorem \ref{thm:optimal_jacobian}, only arises during training if the deep network's Jacobians are regularised to remain bounded. Thus we proposed GrokAlign as a method for doing this.

\section{The Centroid Alignment Perspective}\label{sec:dynmamics_of_deep_network_centroids}

\paragraph{Centroids.} Recall that the functional geometry of a continuous piecewise affine deep network refers to the arrangement of its linear regions. That is, the disjoint union of $\left\{\omega_{\mathbf{x}}\right\}_{\mathbf{x}\in\mathbb{R}^d/\sim}$ where $\sim$ denotes the equivalence class $\mathbf{x}_1\sim\mathbf{x}_2$ if and only if $\mathbf{x}_2\in\omega_{\mathbf{x}_1}$ and vice-versa. Of importance is the fact that the functional geometry can be parametrised with a collection of parameters $\left\{\left(\mu_{\mathbf{x}},\tau_{\mathbf{x}}\right)\right\}_{\mathbf{x}\in\mathbb{R}^d/\sim}\subseteq\mathbb{R}^d\times\mathbb{R}$, termed the {\em centroids} and {\em radii}, according to a power diagram subdivision \citep{balestriero_geometry_2019}.

\begin{theorem}\label{thm:pd_parameters}
    For a continuous piecewise affine deep network, $\mu_{\mathbf{x}}=\left(J_{\mathbf{x}}(f)\right)^\top\mathbf{1}$. (Proof in Appendix \ref{proof:pd_parameters}).
\end{theorem}

Theorem \ref{thm:pd_parameters} tells us that centroids provide a summarisation mechanism for the Jacobian matrices of any deep network in way that has an elegant geometrical interpretation when the network is continuous piecewise affine. Importantly, the centroid can be computed through a Jacobian vector product, which is more computationally efficient than directly computing the Jacobian \citep{balestriero_fast_2021}.

\begin{definition}
    A deep network is \textbf{centroid-aligned} at $\mathbf{x}\in\mathbb{R}^d$ if $\mu_{\mathbf{x}}=c\mathbf{x}$ for some constant $c\in\mathbb{R}$.
\end{definition}

The geometrical consequences of centroid alignment can be visualised vividly in Figure \ref{fig:centroid_alignment}. An aligned centroid can be linked to the \emph{region migration} phenomenon observed in \citet{humayun_deep_2024}, which was used as an explanation for delayed robustness. Region migration describes the process of a deep network migrating its linear regions from the data points to the decision boundary. Therefore, we can already observe that considering centroid alignment will be useful from the perspective of trying to understand the dynamics of deep network training. 

\begin{proposition}\label{prop:jacobian_aligned_implies_centroid_aligned}
    A Jacobian-aligned deep network is centroid-aligned. (Proof in Appendix \ref{proof:jacobian_aligned_implies_centroid_aligned}).
\end{proposition}
From Proposition \ref{prop:jacobian_aligned_implies_centroid_aligned} it follows that we can consider centroid alignment as an alternative to Jacobian alignment. Although centroid alignment is a weaker property than Jacobian alignment, we will demonstrate that it has explicit connections to feature learning through the neural tangent kernel.

\paragraph{Neural Tangent Kernel.} Suppose a deep network has parameters $\theta$. Then the neural tangent kernel \citep{jacot_neural_2018} between $\mathbf{x},\mathbf{x}^\prime\in\mathbb{R}^d$ is taken to be $\Theta\left(\mathbf{x},\mathbf{x}^\prime\right)=\nabla_{\theta}f_{\theta}(\mathbf{x})\left(\nabla_{\theta}f_{\theta}\left(\mathbf{x}^\prime\right)\right)^\top$. The \emph{linear} and \emph{feature} learning regimes of deep network training are characterised by having relatively constant or dynamic neural tangent kernels, respectively \citep{chizat_lazy_2019,woodworth_kernel_2020,moroshko_implicit_2020}. The former identifies when the deep network approximates a linear function, whereas the latter involves the deep network's nonlinearities.\footnote{Lazy and rich are also commonly used terms to refer to these two regimes.}

\paragraph{The Dynamics of Centroids.} For simplicity we will explore the centroid dynamics of a two-layer network of the form $f_{\theta}(\mathbf{x})=W^{(2)}\left(\sigma\left(W^{(1)}\mathbf{x}\right)\right)$, where $W^{(2)}\in\mathbb{R}^{d^{(2)}\times d^{(1)}}$, $W^{(1)}\in\mathbb{R}^{d^{(1)}\times d}$, and $\sigma$ is a piecewise affine nonlinearity (e.g. ReLU). We omit bias terms to ensure the zero bias assumption of Theorem \ref{thm:optimal_jacobian}. We will suppose the deep network is being trained using full-batch gradient descent with a learning rate of $\eta$.

\begin{lemma}\label{lem:centroid_two_layer_network}
    In the setting outlined above, we have $\mu_{\boldsymbol{\mathbf{x}}}=\left(W^{(2)}Q_{\mathbf{x}}W^{(1)}\right)^\top\mathbf{1}$, where $Q_{\mathbf{x}}:=\mathrm{diag}\left(\sigma^\prime\left(W^{(1)}\mathbf{x}\right)\right)$. (Proof in Appendix \ref{proof:centroid_two_layer_network}).
\end{lemma}

To make the connection to feature learning explicit, we will consider the deep network to have a scalar output. However, we provide a treatment of vector-output deep networks in Appendix \ref{sec:vector_output}, where we make an analogous connection between centroids dynamics and feature learning. In the scalar-output setting we suppose that the deep network is being trained with the cross-entropy loss function.


\begin{lemma}\label{lem:ntk}
    In the setting described above, with $d^{(2)}=1$, the neural tangent kernel of $f_{\theta}$ between $\mathbf{x},\mathbf{x}^\prime\in\mathbb{R}^d$ is given by $$\Theta\left(\mathbf{x},\mathbf{x}^\prime\right)=\sigma\left(W^{(1)}\mathbf{x}\right)^\top\sigma\left(W^{(1)}\mathbf{x}^\prime\right)+\left(\mathbf{x}^\top\mathbf{x}^\prime\right)\left(W^{(2)}Q_{\mathbf{x}}Q_{\mathbf{x}^\prime}\left(W^{(2)}\right)^\top\right).$$(Proof in Appendix \ref{proof:ntk}).
\end{lemma}

\begin{theorem}\label{thm:centroid_alignment_ntk}
    In the setting of Lemma \ref{lem:ntk},  we have $\partial_t\left(\left\langle\mathbf{x},\mu_{\mathbf{x}}\right\rangle\right)=\frac{\eta}{m}\sum_{p=1}^m\Theta\left(\mathbf{x},\mathbf{x}_p\right)m_{\mathbf{x}_p}$, where $m_{\mathbf{x}_p}=y_p-\frac{1}{1+\exp\left(-f_{\theta}\left(\mathbf{x}_p\right)\right)}$. (Proof in Appendix \ref{proof:centroid_alignment_ntk}).
\end{theorem}

Theorem \ref{thm:centroid_alignment_ntk} says that the inner product between some point in the input space, $\mathbf{x}$, and its corresponding centroid, $\mu_{\mathbf{x}}$, is a weighted sum of the neural tangent kernel of the point with the points in the training data. In particular, a changing inner product implies a dynamic neural tangent kernel, which corresponds to the feature learning regime of training.

More specifically, if the inner product $\left\langle\mathbf{x},\mu_{\mathbf{x}}\right\rangle$ changes by $\delta$, then the alignment will change by $\frac{\delta}{\Vert\mathbf{x}\Vert\Vert\mu_{\mathbf{x}}\Vert}$. When optimizing a deep network with GrokAlign, we would expect the centroid norm to be low as the centroid is equal to the sum of the rows of the Jacobian (see Theorem \ref{thm:pd_parameters}). Thus, the feature learning regime will be identified by centroid alignment.

Consequently, since Jacobian-aligned deep networks are centroid-aligned and centroid alignment is an indicator of feature learning, we have determined that we can use the degree of centroid alignment as a metric for effectively monitoring the emergence of generalisation and robustness in deep network training dynamics. 
We will now explore this further via a range of numerical experiments.

\section{Experiments}\label{sec:experiments}

To compute the centroid alignment of a deep network, we compute the centroid for an input point using Theorem \ref{thm:pd_parameters} and then compute the cosine similarity between this and the input point. Likewise, we can obtain the centroid inner product. Due to the zero bias assumption of Theorem \ref{thm:optimal_jacobian} and Theorem \ref{thm:optimally_robust}, we will, unless stated otherwise, omit bias terms in the deep networks we consider.

\begin{figure}[t]
    \centering
    \includegraphics[width=\linewidth]{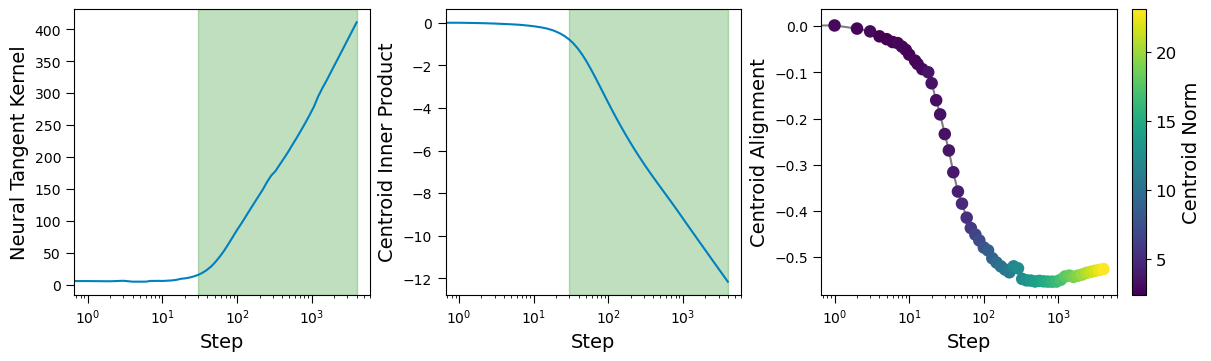}
    \caption{\small Theorem \ref{thm:centroid_alignment_ntk} holds in practice: we observe that a changing inner product indeed corresponds to a feature learning regime. Here we train a two-layer scalar-output ReLU network using the binary-cross-entropy loss function to distinguish between the zero and one class of the MNIST dataset \citep{lecun_gradient-based_1998}. We train the model using full-batch gradient descent for $4000$ steps at a learning rate of $0.01$. At the beginning of training we fix a point from the training set and compute the average value of the neural tangent kernel between itself and the other points from the training set, \textbf{left}. We then compute the centroid of the point using Theorem \ref{thm:pd_parameters} to then obtain the inner product, \textbf{centre}, and its alignment, \textbf{right}. In the right plot we record the norm of the centroid and encode it in the colours of the markers. In the \textbf{left} and \textbf{centre} plots, we identify the feature learning regime using a green shaded area, determined by the neural tangent kernel changing value.}
    \label{fig:ntk}
\end{figure}

\subsection{Centroid Alignment Identifies the Feature Learning Regime}

First, we verify Theorem \ref{thm:centroid_alignment_ntk} and the conclusions stemming from it in an MNIST \citep{lecun_gradient-based_1998} two-class classification setting using a two-layer scalar-output ReLU network. Figure \ref{fig:ntk} demonstrates that the inner product between a point and its linear region's centroid changes in accordance with the neural tangent kernel, meaning that centroid alignment identifies the feature learning regime.

Due to the fact that the norms of the centroids increase during training, eventually the increasing centroid inner product experienced during the feature learning regime no longer translates to centroid alignment. This supports the observation that standard training techniques do not maintain a bounded Jacobian norm \citep{yang_closer_2020}, which highlights the necessity of a method such as GrokAlign to ensure the realisation of a Jacobian-aligned deep network in practice. Without using a method like GrokAlign during training, the initialisation of the network significantly influences its subsequent dynamics. In Appendix  \ref{sec:initialising_for_centroid_alignment} we use centroid alignment to explore this.

\begin{figure}[t]
    \centering
    \includegraphics[width=\linewidth]{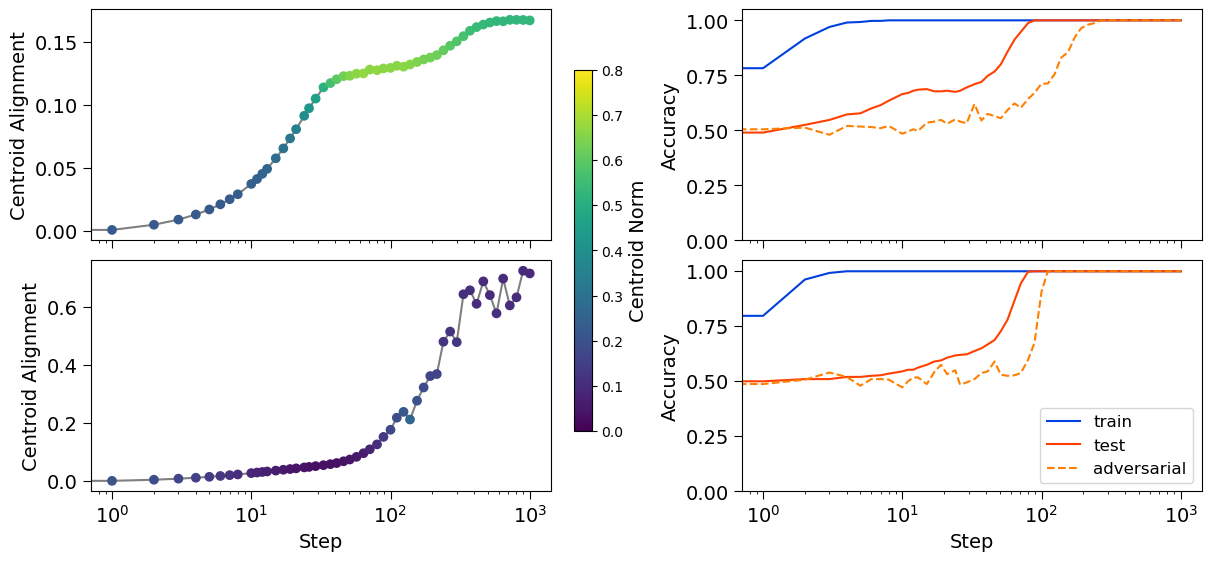}
    \caption{\small Centroid alignment of a point from the training set identifies the generalisation and robustness of a deep network. We study a two-layer network of width $2048$ learning the XOR classification task described in Section \ref{sec:xor_grokking}. In the \textbf{top} row we train the network using full-batch gradient descent under the mean-squared error loss function with a learning rate of $0.1$, weight-decay of $0.1$ and for $1000$ steps. We monitor both the alignment of a training point to its centroid as well as the robustness of the network. To measure robustness, we take the entries from the test set and apply Gaussian perturbations of varying standard deviations to the last $39998$ components. In the \textbf{bottom} row we additionally apply GrokAlign with $\lambda_{\text{Jac}}$ equal to $1.0$. In the \textbf{left} plots we are illustrating the centroid norms with the colours of the markers.}
    \label{fig:xor_grokking}
\end{figure}

\subsection{Centroid Alignment Identifies Delayed Robustness}\label{sec:xor_grokking}

Having demonstrated that centroid alignment identifies the feature learning regime, we now demonstrate that it can be used to identify the onset of robustness.

We adopt a set-up similar to that of \citet{xu2025let} entailing a scalar-output two-layer fully connected network grokking on XOR cluster data. Note that this network trivially has rank-one Jacobians at every point in the input space. The XOR cluster data contains $40000$-dimensional vectors of the form $\mathbf{x}=\left(x_1,x_2,\tilde{\mathbf{x}}^\top\right)^\top\in\mathbb{R}^{40000}$, where $x_1,x_2\in\left\{\pm1\right\}$ and $\tilde{\mathbf{x}}\in\mathbb{R}^{39998}$. The $400$ samples used to train the network are constructed by sampling entries $x_1$, $x_2$ uniformly from $\left\{\pm1\right\}$ and entries of $\tilde{\mathbf{x}}$ uniformly from $\left\{\pm\epsilon\right\}$, here we take $\epsilon=0.05$. The corresponding label of such a sample is $x_1x_2\in\{\pm1\}$. A similar sample is generated as a test set. Therefore, by construction our training data only has \emph{signal} in the first two components, whereas all the other components contain noise. Hence, generalisation would require recognising the pattern of how the first two components lead to the corresponding label, whilst robustness would require the network to not condition its pattern recognition on the last $39998$ components.

Throughout training we track the centroid alignment of a point in the training set to obtain Figure \ref{fig:xor_grokking}. In the top row of Figure \ref{fig:xor_grokking}, we observe that as the network memorises, the centroid alignment does not increase significantly. However, during generalisation, the centroid alignment increases. After a slight plateau in centroid alignment, a further increase correlates with the onset of robustness. In this instance, the rank of the Jacobian of the network at the point under consideration is one, and thus from Theorem \ref{thm:optimally_robust} robustness can only be achieved through alignment. Critically, we observe the indirectness of weight-decay at imposing the Jacobian norm constraint of Theorem \ref{thm:optimal_jacobian}. Early on in training, the norms of the centroids increase and eventually inhibit centroid alignment. Since alignment is an optimum of the training objective (recall Theorem \ref{thm:optimal_jacobian}), it is only at this stage that under weight-decay the network is incentivised to reduce the norms of the Jacobian resulting in the onset of robustness. 

In the bottom row of Figure \ref{fig:xor_grokking}, we instead see that by applying GrokAlign we directly mitigate this delay and achieve robustness much sooner.

\subsection{GrokAlign's Influence on Grokking}\label{sec:controlling_dynamics}

Thus far we have shown that centroid alignment provides a valuable perspective on the training dynamics of a deep network, as it identifies stages of generalisation (see Figure \ref{fig:ntk}) and robustification (see Figure \ref{fig:xor_grokking}). 
We have also shown that standard deep network training, including weight-decay, cannot maintain a Jacobian norm constraint and thus is limited in its ability to Jacobian-align deep networks.

The most direct approach for enforcing the Jacobian norm constraint is through GrokAlign, and here we will explore how this can used to control the deep network's training dynamics that lead to grokking. Recall that GrokAlign regularises the Frobenius norm of the Jacobian matrices of the deep network computed at the training data by appending its value to the loss function. To be computationally efficient, GrokAlign uses an approximation of the Frobenius norm of the Jacobian matrices \citep{hoffman_robust_2019}.\footnote{We provide code for our implementation of GrokAlign \href{https://github.com/ThomasWalker1/grokalign}{here}.}

\begin{figure}[t]
    \centering
    \includegraphics[width=\linewidth]{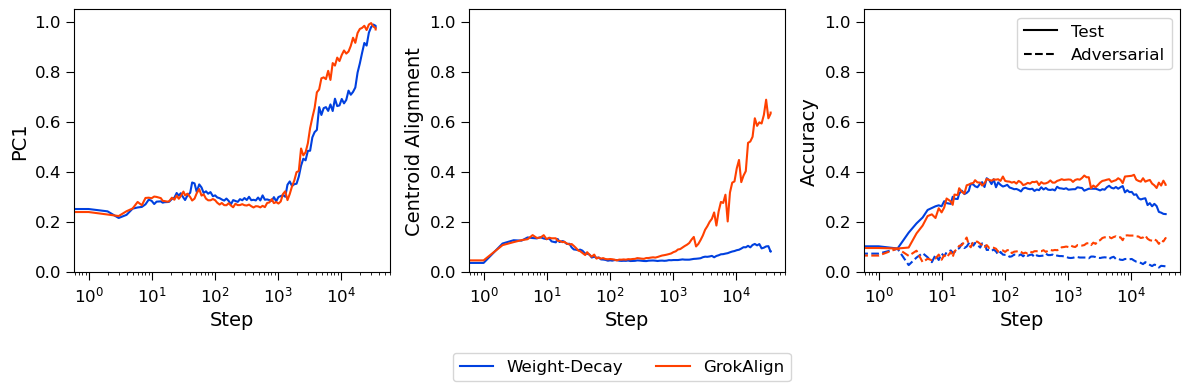}
    \caption{\small GrokAlign capitalises on the low rank implicit bias of deep network training to induce robustness by aligning the Jacobians of the deep network. Here we train a convolutional neural network with six convolutional layers and two linear layers, with no bias terms, on a $1024$ subset of the CIFAR10 dataset \citep{krizhevsky_learning_2009} under the mean-squared error loss function. We use the AdamW optimizer \citep{loshchilov_decoupled_2019} with a learning of $0.001$ and a batch size of $256$ to train the network across $36000$ steps. In one instance we apply weight-decay at $0.001$, and in another instance we apply GrokAlign with $\lambda_{\text{Jac}}$ equal to $0.001$. In the \textbf{left} plot we compute the average explained variance of the first principal component of the Jacobians as in Figure \ref{fig:pc1}, and in the \textbf{centre} plot we record the average centroid alignment on the training set. In the \textbf{right} column we record the test accuracy of the model and the accuracy of the model when $\ell_{\infty}$ perturbations of amplitude $\frac{4}{255}$ are applied to the test set using Autoattack \citep{croce_reliable_2020}.}
    \label{fig:cnn_delayed_robustness}
\end{figure}

\paragraph{Inducing Delayed Robustness.}

We train convolutional neural networks on the CIFAR10 dataset \citep{krizhevsky_learning_2009}. From Figure \ref{fig:cnn_delayed_robustness} we observe that, by using GrokAlign we can induce Jacobian alignment, as evidenced by the increasing centroid alignment, which translates into improved robustness by capitalising on the diminishing ranks of the Jacobian matrices. With only weight-decay, Jacobian alignment does not manifest, as evidenced by the low centroid alignment, and thus we observe a decline in robustness as training progresses. Crucially, by monitoring centroid alignment we can determine that prolonging training is unlikely to improve the properties of the GrokAligned model significantly, since the effective rank of the Jacobians is close to one and the centroid alignments are relatively high and have started plateauing. 

\begin{figure}[h]
     \centering
     \begin{subfigure}{0.32\columnwidth}
         \centering
         \includegraphics[width=0.5\textwidth]{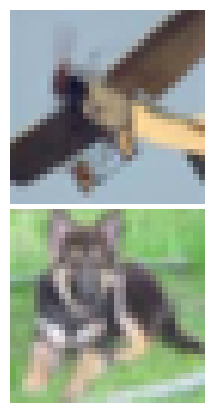}
         \centering
         \caption*{Input}
     \end{subfigure}
     \hfill
     \begin{subfigure}{0.32\columnwidth}
         \centering
         \includegraphics[width=0.5\textwidth]{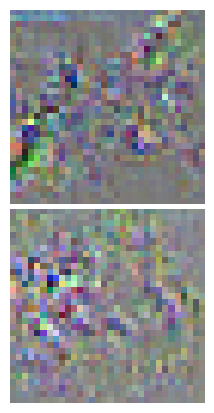}
         \centering
         \caption*{Centroid Unaligned}
     \end{subfigure}
     \hfill
     \begin{subfigure}{0.32\columnwidth}
         \centering
         \includegraphics[width=0.5\textwidth]{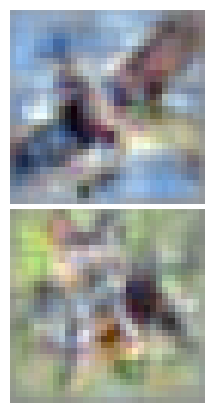}
         \centering
         \caption*{Centroid Aligned}
     \end{subfigure}
     \caption{\small The geometry of the function of a centroid aligned deep network is more resemblant of the geometry of the training data, indicating that the transformation its applies to local regions of data points is more representative of their features. We compare the two convolutional neural networks from Figure \ref{fig:cnn_delayed_robustness} by taking training points, \textbf{left}, and computing their centroid. In the \textbf{centre} we depict the centroids for the network trained without using GrokAlign, and on the \textbf{right} we depict the centroids for the network trained with GrokAlign.}
     \label{fig:cifar_alignment}
\end{figure}

Therefore, we have demonstrated that GrokAlign provides a direct strategy for inducing robustness for it aligns the Jacobian's of the deep networks which we can observe by monitoring centroid alignment. Just like Figure \ref{fig:centroid_alignment}, we can vividly visualise the consequences of centroid alignment in Figure \ref{fig:cifar_alignment}.

\paragraph{Inhibiting Delayed Generalisation.}

Using our reasoning, we would expect that if we were to maintain the Jacobian norms at a high-level, then we ought to prevent alignment and thus generalisation. To explore this, we consider a fully connected network and scale up its weights at initialisation to increase the Jacobian norms, much like \citet{liu_omnigrok_2022}, and apply GrokAlign in different ways.

\begin{figure}[h]
    \centering
    \includegraphics[width=\linewidth]{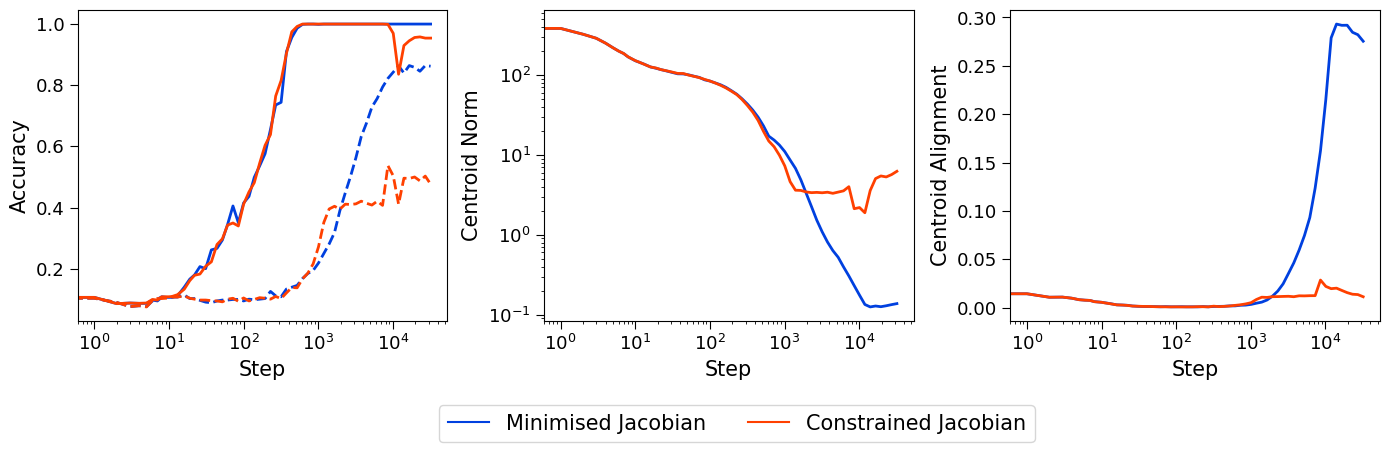}
    \caption{\small By maintaining the Frobenius norm of the Jacobians at the training data relatively high we can keep the norms of the centroids relatively high which prevents grokking. We take the MNIST grokking set up of \citet{liu_omnigrok_2022}. In the minimising case we impose GrokAlign with $\lambda_{\text{Jac}}$ equal to $0.001$ during training to minimise the Frobenius norm of the Jacobians at the training data. In the constrained case we apply GrokAlign to maintain the Frobenius norm of the Jacobians computed at the training data at a relatively high level. More specifically, using a regularisation coefficient of $0.001$, we append the difference between five and the average Frobenius norm of the training data to the loss function. In the \textbf{left} plot we visualise the train and test accuracy with solid and dashed lines respectively. In the \textbf{centre} plot we visualise the average norm of the centroids computed at the training data. In the \textbf{right} plot we visualise the average centroid alignment.}
    \label{fig:restricted_jacobian}
\end{figure}

We observe in Figure \ref{fig:restricted_jacobian} that our prediction is correct: minimising the Frobenius norms of the Jacobians leads to generalisation, whilst keeping their value relatively high prevents it. We are able to monitor this by tracking the norms of the centroids, which demonstrates how the centroids provide an effective mechanism to monitor network dynamics.

\paragraph{Accelerating Grokking.}

Just as we used GrokAlign to inhibit grokking, we can also use it to accelerate grokking. For this we consider the standard MNIST grokking set up of \citet{liu_omnigrok_2022} which involves applying weight-decay to a deep network initialised with a large initial weight-norm. From our perspective this results in Jacobians with large norm, inhibiting their alignment.

We compare the effectiveness of GrokAlign to two other known methods of inducing grokking. Grokfast \citep{lee_grokfast_2024} works to improve the rate of grokking by manipulating the gradients during training to amplify certain signals. \citet{tan_understanding_2024} motivated adversarial training for accelerating grokking by establishing a connection between robustness and generalisation. The method of adversarial training involves perturbing the inputs during training with noise proportional to the training accuracy of the deep network. In each method we do not manipulate the weight-decay of the training procedure.

From Table \ref{tab:jr_grokking_improvements}, we observe that GrokAlign is extremely effective at inducing the grokked state of the deep network in this setting; it arrives at the grokked state in $7.56$ times fewer steps and $6.31$ times faster than the baseline in the case of the cross-entropy loss function. In contrast, Grokfast provides a relatively lower improvement in the mean-squared error case and is ineffective in the cross-entropy case. Adversarial training does not improve the rate of grokking over the baseline. In particular, adversarial training does not improve the robustness of the model by way of aligning the Jacobian, unlike GrokAlign (see Appendix \ref{sec:adversarial_training_alignment}).

\begin{table}[h]
     \caption{\small GrokAlign significantly speeds up the rate of grokking. In the \textbf{top} table we assume the MNIST set up of \citet{liu_omnigrok_2022} with the cross-entropy loss function, and in the \textbf{bottom} table we assume it with mean-squared error loss function as the baselines. We repeat the training across ten different random initialisations, where for the cross-entropy loss function we apply GrokAlign with $\lambda_{\text{Jac}}$ equal to $0.001$ and with $\lambda_{\text{Jac}}$ equal to $0.0001$ for the mean-squared error models. For each run we measure the number of steps and the absolute time, in seconds, taken for the networks to reach $85$\% test accuracy. In the case of the mean-squared error loss function, we additionally measure the time, in seconds, for the models to go from $20$\% to $85$\% test accuracy. We provide the average acceleration (or deceleration) of each method compared to the baseline along with the corresponding standard deviation.}
     \label{tab:jr_grokking_improvements}
     \vspace{1em}
    \begin{subtable}[h]{\textwidth}
        \centering
        \caption*{Cross Entropy Loss Function}
        \begin{tabular}{lcc}
        \midrule
        Regularisation & Number of Steps & Absolute Time (s) \\
        \midrule
        Baseline & -- & -- \\
        GrokAlign & $\boldsymbol{\downarrow}\mathbf{7.56}\boldsymbol{\times}(\pm0.82)$ & $\boldsymbol{\downarrow}\mathbf{6.41}\boldsymbol{\times}(\pm0.69)$ \\
        Grokfast & $\uparrow 1.01\times(\pm0.04)$ & $\uparrow 1.03\times(\pm0.04)$ \\
        Adversarial Training & $\downarrow 1.32\times(\pm0.18)$ & $\uparrow 1.92\times(\pm0.29)$ \\
        \midrule
        \end{tabular}
    \end{subtable}
    \begin{subtable}[h]{\textwidth}
        \centering
        \caption*{Mean Squared Error Loss Function}
        \begin{tabular}{lccc}
        \midrule
        Regularisation & Number of Steps & Absolute Time (s) & Grokking Phase Time (s) \\
        \midrule
        Baseline & -- & -- & -- \\
        GrokAlign & $\boldsymbol{\downarrow}\mathbf{1.69}\boldsymbol{\times}(\pm0.31)$ & $\boldsymbol{\downarrow}\mathbf{1.46}\boldsymbol{\times}(\pm0.30)$ & $\boldsymbol{\downarrow}\mathbf{1.77}\boldsymbol{\times}(\pm0.53)$ \\
        Grokfast & $\downarrow 1.08\times(\pm0.17)$ & $\downarrow 1.05\times(\pm0.23)$ & $\downarrow 1.05\times(\pm0.27)$ \\
        Adversarial Training & $\downarrow 1.23\times(\pm0.26)$ & $\uparrow 1.92\times(\pm0.29)$ & $\uparrow 2.02\times(\pm0.37)$ \\
        \midrule
        \end{tabular}
     \end{subtable}
\end{table}

\paragraph{Controlling the Learned Solutions of Deep Networks.}\label{sec:algorithmic}

The computations of centroids is valid without the continuous piecewise affine assumption; it is only their interpretation as characterising a functional geometry that requires the assumption. Therefore, we can examine the centroid alignment of more elaborate models like transformers \citep{vaswani_attention_2017}.

\citet{nanda_progress_2022} observed that a single layer transformer learning modular addition \citep{power_grokking_2022} grokked by learning how to implement an algorithm. An equally viable solution to this problem would be through classification. Since our framework is largely motivated in the classification setting, we can explore the tension between these solutions through centroid alignment.

\begin{figure}[h]
    \centering
    \includegraphics[width=\linewidth]{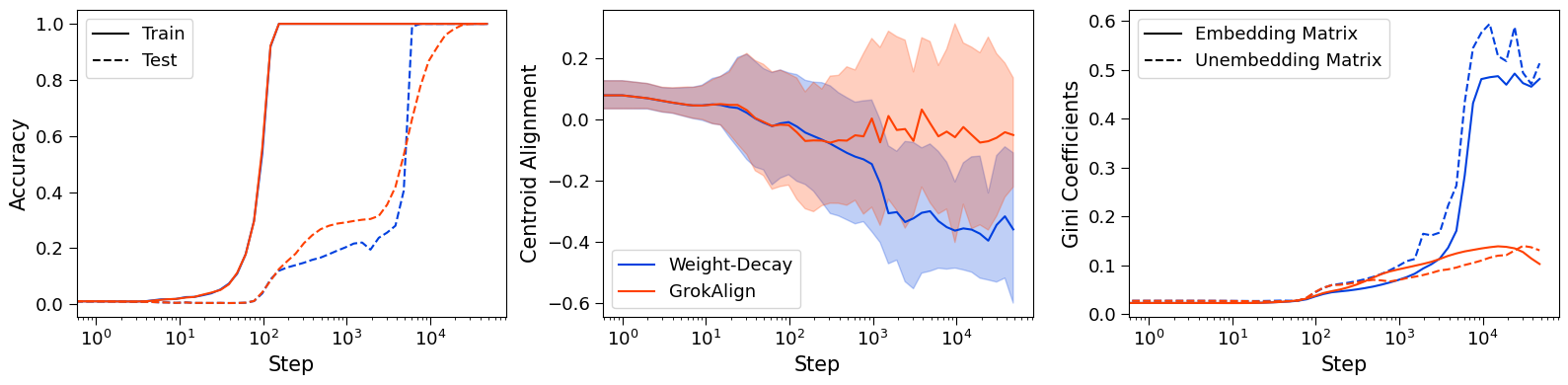}
    \caption{\small GrokAlign biases a single layer transformer model to learn a \emph{classification} style solution. Here we obtain the centroid alignment statistics for a single layer transformer trained on modular addition \citep{power_grokking_2022}. We use the same training pipeline as in \citet{nanda_progress_2022}, with the added utilisation of GrokAlign with $\lambda_{\text{Jac}}$ equal to $0.001$. In the \textbf{left} plot we record the accuracies of the transformer model on the train and test set. In the \textbf{centre} plot we record the centroid alignment of the transformer model from the continuous embeddings of the discrete input tokens to the output of the model; the shaded region illustrates the maximum and minimum alignment observed with the solid line corresponding to the mean. In the \textbf{right} plot we record the Gini coefficients of the embedding and unembedding matrices of the model as proposed in \citet{nanda_progress_2022}.}
    \label{fig:transformer_alignment_learnable}
\end{figure}

In Figure \ref{fig:transformer_alignment_learnable} we consider the centroid alignment of the transformer model under the standard training pipeline of \citet{nanda_progress_2022} with the added utilisation of GrokAlign. As with our previous experiments, we observe that centroid alignment changes in accordance with test accuracy. However, it is not as salient as in previous cases, and applying GrokAlign does not accelerate the rate of grokking. This is perhaps due to the model learning the algorithmic solution to the task, which is not entirely compatible with the Jacobian and centroid aligned perspective.

We support the idea that GrokAlign biases toward the classification style solution by tracking the Gini coefficient of the embedding and unembedding matrices \citep{nanda_progress_2022}. A key aspect that allows the transformer to implement its algorithmic solution is the ability to manipulate the embedding and unembedding matrices \citep{nanda_progress_2022}. In particular, the Gini coefficient of these embedding matrices should be relatively higher when implementing the algorithmic solution. We observe that when applying GrokAlign, these Gini coefficient remain low, indicating that GrokAlign encourages the model to learn the classification style solution which is not the natural solution solution in this particular training regime (see the right plot of Figure \ref{fig:transformer_alignment_learnable}). Consequently, we observe its ineffectiveness in accelerating grokking. 

\begin{figure}[h]
    \centering
    \includegraphics[width=\linewidth]{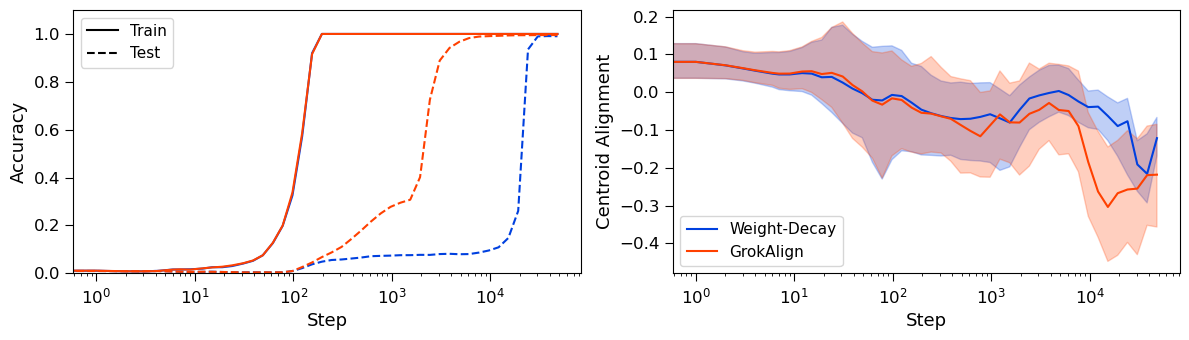}
    \caption{\small Without the ability to manipulate its embedding matrices, a single layer transformer learns the classification style solution for performing modular addition and thus benefits from being trained with GrokAlign. We adopt the same configurations as in Figure \ref{fig:transformer_alignment_learnable}, except we fix the embedding and unembedding matrices of the model during training.}
    \label{fig:transformer_alignment_fixed}
\end{figure}

If we instead fix the embedding matrix during training, we a priori bias the the model to learn the classification style solution. Under this set up (see Figure \ref{fig:transformer_alignment_fixed}), we observe that the transformer groks earlier with GrokAlign than with weight-decay.

\section{Discussion}\label{sec:discussion}

\paragraph{Impacts.} We have identified that Jacobian alignment is the cause for grokking, by demonstrating theoretically and empirically that Jacobian-aligned deep networks optimise the loss function under a Jacobian norm constraint and are optimally robust under the low rank bias of training dynamics. Consequently, we identified regularising the norms of the Jacobians of a deep network as an effective strategy for controlling the dynamics of deep networks to instil particular properties -- a method we introduce as GrokAlign. In particular, we showed that using GrokAlign we can induce robustness, inhibit or accelerate grokking and dictate the learned solutions of deep networks. 

Since Jacobian matrices are difficult to interpret and costly work with in practice, we constructed the centroid alignment perspective as an alternative strategy to monitor the dynamics of deep networks. This perspective is interpretable due to its relationship with the functional geometry of a deep network and is theoretically meaningful due to its connection to the neural tangent kernel. Using this perspective we were able to identify the onset of generalisation and robustness during deep network training, as well as reason about when prolonging training would improve these important properties.

\paragraph{Limitations.}
We have developed our theory in the context of continuous piecewise affine networks for convenience, although its implications extend to any deep network architecture, including transformers. Our experiments have mainly considered continuous piecewise networks. Understanding how our perspective holds outside this setting is perhaps warranted. We provide initial results on a transformer in Section \ref{sec:algorithmic}.

Furthermore, some of our reasoning is dependent on the assumption that deep network training dynamics minimise the rank of the weight matrices. Although this has been theoretically demonstrated in particular settings and observed empirically in practice, a deeper characterisation of this phenomenon is necessary.

\ack{This work was supported by ONR grant N00014-23-1-2714, ONR MURI N00014-20-1-2787, DOE grant DE-SC0020345, and DOI grant 140D0423C0076.}

\newpage
\bibliography{references.bib}
\bibliographystyle{unsrtnat}


\newpage
\appendix


\section{Centroid Dynamics of Vector-Output Deep Networks}\label{sec:vector_output}

Consider the case of a general vector output, namely $d^{(2)}\geq2$, with $\mathcal{L}$ being the cross-entropy loss function or the mean-squared error loss function. Namely, we consider $$\ell\left(f\left(\mathbf{x}_p\right),y_p\right)=-\log\left(\frac{\exp\left(\left[f\left(\mathbf{x}_p\right)\right]_{y_p}\right)}{\sum_{c=1}^{d^{(2)}}\exp\left(\left[f\left(\mathbf{x}_p\right)\right]_{c}\right)}\right)$$for the cross-entropy loss function, or $$\ell\left(f\left(\mathbf{x}_p\right),y_p\right)=\left\Vert\mathbf{e}_{y_p}-f\left(\mathbf{x}_p\right)\right\Vert_2^2$$for the mean-squared error loss function.

\begin{proposition}\label{prop:centroid_dynamics}
    In the setting described above, we have 
    \begin{align*}
        \partial_t\left(\left\langle\mathbf{x},\mu_{\mathbf{x}}\right)\right\rangle&=\frac{\eta}{m}\sum_{p=1}^m\Bigg(\left(\mathbf{m}_{\mathbf{x}_p}^\top W^{(2)}Q_{\mathbf{x}_p}Q_{\mathbf{x}}\left(W^{(2)}\right)^\top\mathbf{1}\right)\left\langle\mathbf{x},\mathbf{x}_p\right\rangle\\&\quad\qquad+\mathbf{x}^\top\left(W^{(1)}\right)^\top Q_{\mathbf{x}}\sigma\left(W^{(1)}\mathbf{x}_p\right)\mathbf{m}_{\mathbf{x}_p}^\top\mathbf{1}\Bigg)
    \end{align*}
    where$$\mathbf{m}_{\mathbf{x}_p}=\mathbf{e}_y-\frac{\exp\left(\left[f_{\theta}\left(\mathbf{x}_p\right)\right]_{y_p}\right)}{\sum_{c=1}^C\exp\left(\left[f_{\theta}\left(\mathbf{x}_p\right)\right]_c\right)}$$in the case of the cross-entropy loss function and $$\mathbf{m}_{\mathbf{x}_p}=2\left(\mathbf{e}_y-f_{\theta}\left(\mathbf{x}_p\right)\right).$$
\end{proposition}

\begin{corollary}
    In the setting of Proposition \ref{prop:centroid_dynamics}, under the cross entropy loss function, we have 
    \begin{align*}
        \partial_t\left(\left\langle\mathbf{x},\mu_{\mathbf{x}}\right)\right\rangle&=\frac{\eta}{m}\sum_{p=1}^m\left(\mathbf{m}_{\mathbf{x}_p}^\top W^{(2)}Q_{\mathbf{x}_p}Q_{\mathbf{x}}\left(W^{(2)}\right)^\top\mathbf{1}\right)\left\langle\mathbf{x},\mathbf{x}_p\right\rangle\\&:=\frac{\eta}{m}\sum_{p=1}^m\frac{\iota_{\mathbf{x},p}}{\left\Vert\mathbf{x}_p\right\Vert_2}\left\langle\mathbf{x},\mathbf{x}_p\right\rangle
    \end{align*}
\end{corollary}

That is, in the context of the cross-entropy loss function, the centroids are aligned in a manner that is proportional to the alignment of their encompassing points with the training data. More specifically, with the intuition that the role of $W^{(2)}$ in $f_{\theta}$ is to be a collection of filters facilitating the classification of each class, the quantity $W^{(2)}Q_{\mathbf{x}_p}\left(W^{(2)}Q_{\boldsymbol{\nu}}\right)^\top\mathbf{1}$ can be thought of as identifying how each feature of $\mathbf{x}_p$ correlates with the features of the region $\omega_{\boldsymbol{\nu}}$. Since $\mathbf{m}\left[\mathbf{x}_p\right]$ is positive on the correct class and negative for the incorrect classes, the term $\iota_{\mathbf{x}^\prime,p}$ is largest when $\omega_{\mathbf{x}^\prime}$ has identified features that correlate with the features of $\mathbf{x}_p$ that indicate the class it belongs too. In such a case, the centroid $\mu_{\mathbf{x}^\prime}$ moves in the direction of $\mathbf{x}_p$ to further maximize this correlation. Showing how regions $\omega_{\mathbf{x}^\prime}$ are being optimized to capture the features of classes that help it distinguish it from the other classes. Therefore, we can see neural network training more as a process of allocating linear regions to different features that best distinguishes themselves from the other classes. This would suggest that when we observe the centroids of a layer of a neural network aligning with the data it encompasses, the neural network is performing feature extraction. In particular, the centroid of a training point is incentivised to positively align with itself.

\section{Initialising for Centroid Alignment}\label{sec:initialising_for_centroid_alignment}

When regularisation is not used to train a deep network, the initialisation of the deep network parameters plays a significant role in determining the extent to which is centroids align. In particular, there is a tension between increasing the rate of change of the centroid inner product and maintaining a low centroid norm to translate the increasing inner product to an increase in alignment.

For example, increasing the rate of change of inner product can be done by increasing the neural tangent kernel, say through scaling the weights or output of the network. However, these will also increase the norm of the centroid. Moreover, such scaling is known to increase the propensity of the network maintaining a linear learning regime \citep{chizat_lazy_2019}, along with increasing the width of the neural network \citep{lee_wide_2019} and label rescaling \citep{geiger_disentangling_2020}. We explore this trade-off by repeating the experiment of Figure \ref{fig:ntk}, but with various scaling of the weights or output of the network.

\begin{figure}[h]
    \centering
    \includegraphics[width=\linewidth]{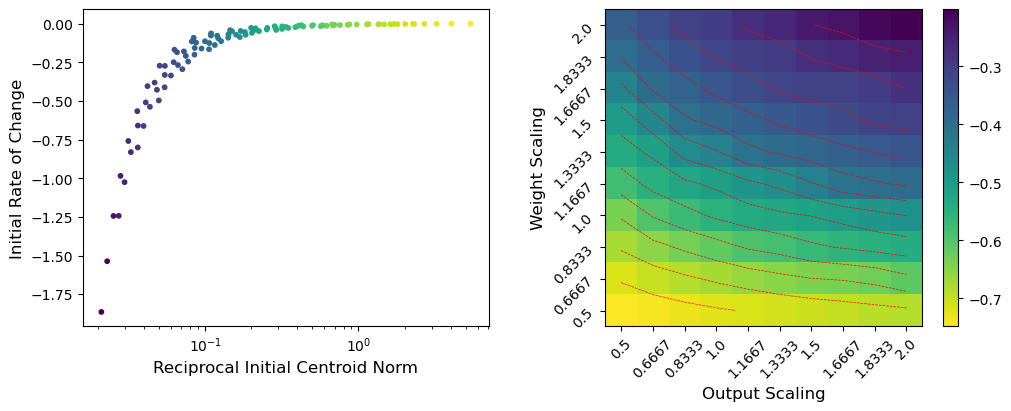}
    \caption{\small When no regularisation is used, minimising the centroid norm at initialisation is essential for ensuring the alignment of the centroid during training, and output scaling ensures this more effectively than scaling the weights at initialisation. Here we repeat the experiment of Figure \ref{fig:ntk}, but with varying scaling of the weights and output of the network. On the \textbf{left} we plot the correlation between the initial rate of change of inner product, as computed by Theorem \ref{thm:centroid_alignment_ntk}, and the average reciprocal of the norm of the centroids of the training points. We additionally colour the scatter points according to the maximum alignment of the centroid observed during training. On the \textbf{right} we observe how the maximum alignment of the centroid observed during training correlates with the different scaling mechanisms.}
    \label{fig:scaling}
\end{figure}

In Figure \ref{fig:scaling}, we observe that controlling the norm of the centroid is a more effective strategy for translating the feature learning of the deep network into centroid alignment. However, a priori, knowing how to initialise the deep network for favourable alignment dynamics is challenging, hence, in practice some sort of regularisation is necessary.

\section{Alignment Induced by Adversarial Training}\label{sec:adversarial_training_alignment}

Although both GrokAlign and adversarial training are motivated to induce grokking by improving the model's robustness, the former does this though aligning the functional geometry of the model, whereas the latter does not. We determine this by measuring the cosine similarity between training points and the rows of the Jacobian of the model at those points (see Table \ref{tab:adversarial_training}).

\begin{table}[h]
     \caption{\small We compare adversarial training to the baseline and GrokAlign grokking set ups of Table \ref{tab:jr_grokking_improvements} in terms of inducing the alignment of the Jacobian at the training data.}
     \vspace{1em}
    \begin{subtable}[h]{\textwidth}
        \centerline{\begin{tabular}{lcccc}
        \midrule
        Regularisation & Test Accuracy & Jacobian Row Alignment (max/min) \\
        \midrule
        Baseline & $88.9$\% & $-0.254$/$0.283$ \\
        \midrule
        GrokAlign & $\mathbf{91.8}$\% & $\mathbf{-0.509}$/$\mathbf{0.478}$ \\
        \midrule
        Adversarial Training & $88.0$\% & $-0.253$/$0.274$ \\
        \midrule
        \end{tabular}}
    \end{subtable}
    \hfill
    \begin{subtable}[h]{\textwidth}
        \centerline{
        \begin{tabular}{lcccc}
        \midrule
        Regularisation & Test Accuracy & Jacobian Row Alignment (max/min) \\
        \midrule
        Baseline & $86.8$\% & $-0.170$/$0.272$ \\
        \midrule
        GrokAlign & $\mathbf{88.2}$\% & $\mathbf{-0.638}$/$\mathbf{0.709}$ \\
        \midrule
        Adversarial Training & $87.8$\% & $-0.263$/$0.376$ \\
        \midrule
        \end{tabular}}
     \end{subtable}
     \label{tab:adversarial_training}
\end{table}

\section{The Geometry of Centroid Alignment}

Centroids have a known connection to the functional geometry of deep networks through the spline theory of deep learning \citep{balestriero_spline_2018}. They form part of the parameter of a region in the power diagram subdivision formulation of deep network's functional geometry \citep{balestriero_geometry_2019}. In Figure \ref{fig:centroid_alignment} we demonstrated the implications of centroid alignment on this geometry. More specifically, we showed that it corresponds to the linear regions flattening around inputs of the same class, in a similar way to the region migration phenomenon identified in \citet{humayun_deep_2024}. Here we support that claim be illustrating it beyond the input points considered in Figure \ref{fig:centroid_alignment}.

\begin{figure}[h]
  \centering
  \begin{subfigure}{0.3\columnwidth}
    \includegraphics[width=0.9\textwidth]{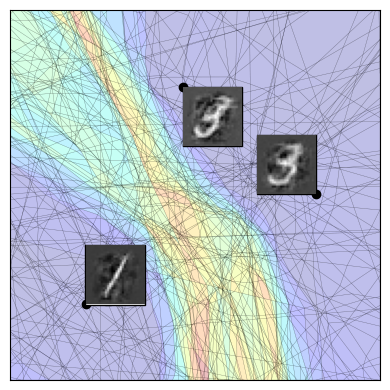}
 \end{subfigure}
  \hspace{0.5em}
  \begin{subfigure}{0.3\columnwidth}
    \includegraphics[width=0.9\textwidth]{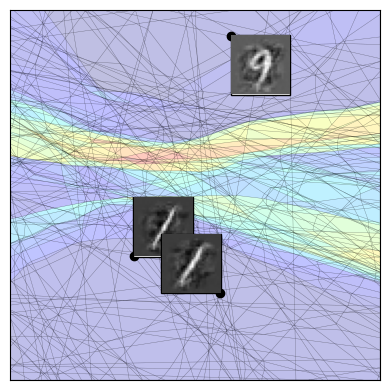}
 \end{subfigure}
  \hspace{0.5em}
  \begin{subfigure}{0.3\columnwidth}
    \includegraphics[width=0.9\textwidth]{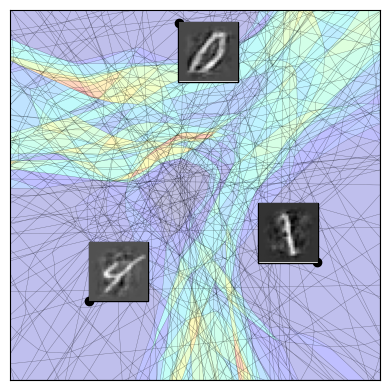}
 \end{subfigure}
  \begin{subfigure}{0.3\columnwidth}
    \includegraphics[width=0.9\textwidth]{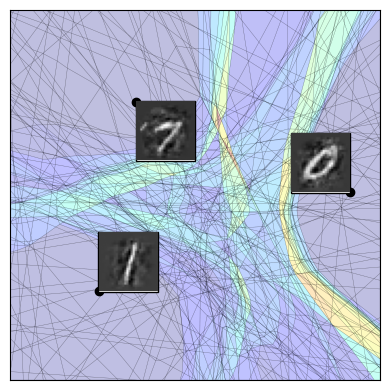}
 \end{subfigure}
  \hspace{0.5em}
  \begin{subfigure}{0.3\columnwidth}
    \includegraphics[width=0.9\textwidth]{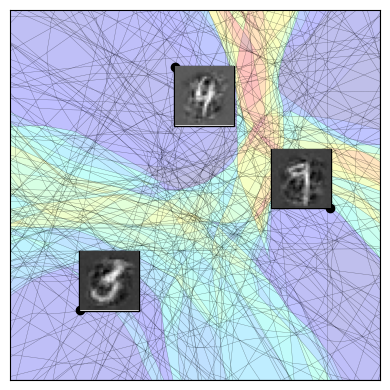}
 \end{subfigure}
  \hspace{0.5em}
  \begin{subfigure}{0.3\columnwidth}
    \includegraphics[width=0.9\textwidth]{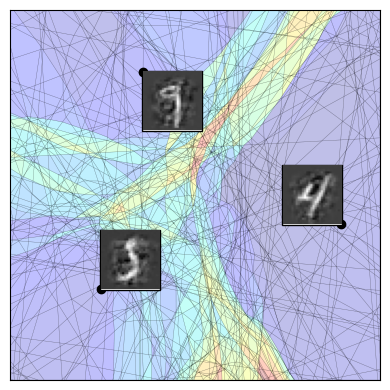}
 \end{subfigure}
  \caption{\small Here we corroborate the right plot of Figure \ref{fig:centroid_alignment} by replicating it on other samples in the input space.}
  \label{fig:centroid_alignment2}
\end{figure}

\vspace{4em}

\section{Proofs of Main Results}\label{sec:proofs}

\paragraph{Proof of Theorem \ref{thm:optimal_jacobian}.}\label{proof:optimal_jacobian}

\begin{enumerate}
    \item In the instance of the cross-entropy loss function, $$\ell_p:=\ell\left(f\left(\mathbf{x}_p\right),y_p\right)=-\log\left(\frac{\exp\left(\left[f\left(\mathbf{x}_p\right)\right]_{y_p}\right)}{\sum_{c=1}^C\exp\left(\left[f\left(\mathbf{x}_p\right)\right]_c\right)}\right).$$Under the assumptions, the output of the neural network at $\mathbf{x}_p$ is $A_{\omega_{\mathbf{x}_p}}\mathbf{x}_p$. The cross entropy loss of the deep network on $\mathcal{D}$ is $$\mathcal{L}_{\text{CE}}=\frac{1}{m}\sum_{p=1}^m\ell_p$$where
    \begin{align*}
        \ell_p&=-\log\left(\frac{\exp\left(\left[A_{\omega_{\mathbf{x}_p}}\mathbf{x}_p\right]_{y_p,\cdot}\right)}{\sum_{c=1}^C\exp\left(\left[A_{\omega_{\mathbf{x}_p}}\mathbf{x}_p\right]_{c,\cdot}\right)}\right)\\&=-\left\langle\left[A_{\omega_{\mathbf{x}_p}}\right]_{y_p,\cdot},\mathbf{x}_p\right\rangle+\log\left(\sum_{c=1}^C\exp\left(\left\langle\left[A_{\omega_{\mathbf{x}_p}}\right]_{c,\cdot},\mathbf{x}_p\right\rangle\right)\right),
    \end{align*}
    which is convex on a convex set. Thus we can consider the sufficient Karush-Kuhn-Tucker conditions with Lagrange multiplier, $$\mathcal{L}=\mathcal{L}_{\text{CE}}+\lambda\left(\sum_{c=1}^C\left\Vert\left[A_{\omega_{\mathbf{x}_p}}\right]_{c,\cdot}\right\Vert_2^2-\alpha\right).$$In particular, the Karush-Kuhn-Tucker conditions have the form
    \begin{align}\label{eq:first_kkt_conditions_ce}
        \frac{\partial\mathcal{L}}{\partial\left[A_{\omega_{\mathbf{x}_p}}\right]_{c,\cdot}}&=-\mathbf{1}_{\left\{y_p=c\right\}}\mathbf{x}_p+\frac{\mathbf{x}_p\exp\left(\left\langle\left[A_{\omega_{\mathbf{x}_p}}\right]_{c,\cdot},\mathbf{x}_p\right\rangle\right)}{\sum_{c^\prime=1}^C\exp\left(\left\langle\left[A_{\omega_{\mathbf{x}_p}}\right]_{c^\prime,\cdot},\mathbf{x}_p\right\rangle\right)}-2\lambda\left[A_{\omega_{\mathbf{x}_p}}\right]_{c,\cdot}\nonumber\\&=\mathbf{0},
    \end{align}
    for $c=1,\dots,C$ and 
    \begin{equation}\label{eq:second_kkt_condition_ce}
        \frac{\partial\mathcal{L}}{\partial\lambda}=\alpha-\sum_{c=1}^C\left\Vert\left[A_{\omega_{\mathbf{x}_p}}\right]_{c,\cdot}\right\Vert_2^2=0.
    \end{equation}
    From \eqref{eq:first_kkt_conditions_ce}, we have
    \begin{align*}
        0&=\sum_{c=1}^C\left\langle\left[A_{\omega_{\mathbf{x}_p}}\right]_{c,\cdot},\frac{\partial\mathcal{L}}{\partial\left[A_{\omega_{\mathbf{x}_p}}\right]_{c,\cdot}}\right\rangle\\&=-\left\langle\left[A_{\omega_{\mathbf{x}_p}}\right]_{y_p,\cdot},\mathbf{x}_p\right\rangle+\sum_{c=1}^C\frac{\left\langle\left[A_{\omega_{\mathbf{x}_p}}\right]_{c,\cdot},\mathbf{x}_p\right\rangle\exp\left(\left\langle\left[A_{\omega_{\mathbf{x}_p}}\right]_{c,\cdot},\mathbf{x}_p\right\rangle\right)}{\sum_{c^\prime=1}^C\exp\left(\left\langle\left[A_{\omega_{\mathbf{x}_p}}\right]_{c^\prime,\cdot},\mathbf{x}_p\right\rangle\right)}\\&\quad-2\lambda\sum_{c=1}^C\left\Vert\left[A_{\omega_{\mathbf{x}_p}}\right]_{c,\cdot}\right\Vert_2^2.
    \end{align*}
    Let $\varrho_c=\frac{\exp\left(\left\langle\left[A_{\omega_{\mathbf{x}_p}}\right]_{c,\cdot},\mathbf{x}_p\right\rangle\right)}{\sum_{c^\prime=1}^C\exp\left(\left\langle\left[A_{\omega_{\mathbf{x}_p}}\right]_{c^\prime,\cdot},\mathbf{x}_p\right\rangle\right)}$. Then, in conjunction with \eqref{eq:second_kkt_condition_ce}, it follows that $$\lambda=\frac{1}{2\alpha}\sum_{c=1}^C\left\langle\left[A_{\omega_{\mathbf{x}_p}}\right]_{c,\cdot},\mathbf{x}_p\right\rangle\varrho_c-\left\langle\left[A_{\omega_{\mathbf{x}_p}}\right]_{y_p,\cdot},\mathbf{x}_p\right\rangle.$$Using this back in \eqref{eq:first_kkt_conditions_ce} we obtain
    \begin{align*}
        \frac{\partial\mathcal{L}}{\partial\left[A_{\omega_{\mathbf{x}_p}}\right]_{c,\cdot}}=&-\mathbf{1}_{\left\{y_p=c\right\}}\mathbf{x}_p+\mathbf{x}_p\varrho_c+\frac{1}{\alpha}\left\langle\left[A_{\omega_{\mathbf{x}_p}}\right]_{y_p,\cdot},\mathbf{x}_p\right\rangle\left[A_{\omega_{\mathbf{x}_p}}\right]_{c,\cdot}\\&-\frac{1}{\alpha}\sum_{c^\prime=1}^C\varrho_{c^\prime}\left\langle\left[A_{\omega_{\mathbf{x}_p}}\right]_{c^\prime,\cdot},\mathbf{x}_p\right\rangle\left[A_{\omega_{\mathbf{x}_p}}\right]_{c,\cdot}\\=&\left(-\mathbf{1}_{\left\{y_p=c\right\}}+\varrho_c\right)\mathbf{x}_p+\frac{\left\langle\left[A_{\omega_{\mathbf{x}_p}}\right]_{y_p,\cdot},\mathbf{x}_p\right\rangle\left[A_{\omega_{\mathbf{x}_p}}\right]_{c,\cdot}}{\alpha}\left(1-\varrho_{y_p}\right)\\&-\frac{1}{\alpha}\sum_{c^\prime\neq y_p}\left\langle\left[A_{\omega_{\mathbf{x}_p}}\right]_{c^\prime,\cdot},\mathbf{x}_p\right\rangle\left[A_{\omega_{\mathbf{x}_p}}\right]_{c,\cdot}\varrho_{c^\prime}\\=&\left(-\mathbf{1}_{\left\{y_p=c\right\}}+\varrho_c\right)\mathbf{x}_p+\frac{\left\langle\left[A_{\omega_{\mathbf{x}_p}}\right]_{y_p,\cdot},\mathbf{x}_p\right\rangle\left[A_{\omega_{\mathbf{x}_p}}\right]_{c,\cdot}}{\alpha}\left(1-\varrho_{y_p}\right)\\&-\frac{1}{\alpha}\left\langle\left[A_{\omega_{\mathbf{x}_p}}\right]_{i,\cdot},\mathbf{x}_p\right\rangle\left[A_{\omega_{\mathbf{x}_p}}\right]_{c,\cdot}\left(1-\varrho_{y_p}\right)\\=&\left(-\mathbf{1}_{\left\{y_p=c\right\}}+\varrho_c\right)\mathbf{x}_p\\&+\frac{\left(1-\varrho_{y_p}\right)\left[A_{\omega_{\mathbf{x}_p}}\right]_{c,\cdot}}{\alpha}\left(\left\langle\left[A_{\omega_{\mathbf{x}_p}}\right]_{y_p,\cdot},\mathbf{x}_p\right\rangle-\left\langle\left[A_{\omega_{\mathbf{x}_p}}\right]_{i,\cdot},\mathbf{x}_p\right\rangle\right),
    \end{align*}
    where $i$ is just some incorrect class for $\mathbf{x}_p$, namely $i\neq y_p$. When $c=y_p$ this reduces to $$\frac{\partial\mathcal{L}}{\partial\left[A_{\omega_{\mathbf{x}_p}}\right]_{c,\cdot}}=\left(1-\varrho_{y_p}\right)\left(-\mathbf{x}_p+\frac{1}{\alpha}\left\langle\left[A_{\omega_{\mathbf{x}_p}}\right]_{y_p,\cdot}-\left[A_{\omega_{\mathbf{x}_p}}\right]_{i,\cdot},\mathbf{x}_p\right\rangle\left[A_{\omega_{\mathbf{x}_p}}\right]_{y_p,\cdot}\right),$$and when $c\neq y_p$ it reduces to $$\frac{\partial\mathcal{L}}{\partial\left[A_{\omega_{\mathbf{x}_p}}\right]_{c,\cdot}}=\left(1-\varrho_{y_p}\right)\left(\frac{1}{C-1}\mathbf{x}_p+\frac{1}{\alpha}\left\langle\left[A_{\omega_{\mathbf{x}_p}}\right]_{y_p,\cdot}-\left[A_{\omega_{\mathbf{x}_p}}\right]_{i,\cdot},\mathbf{x}_p\right\rangle\left[A_{\omega_{\mathbf{x}_p}}\right]_{c,\cdot}\right).$$To obtain the optimal $A$, it suffices to find $A_{\omega_{\mathbf{x}_p}}$ satisfying these conditions. To do so we consider the ansatz $$\left[A_{\omega_{\mathbf{x}_p}}\right]_{c,\cdot}=\begin{cases}a\mathbf{x}_p&c=y_p\\b\mathbf{x}_p&c\neq y_p.\end{cases}$$Substituting this into our conditions we obtain the equations $$\begin{cases}-1+\frac{1}{\alpha}(a-b)a\left\Vert\mathbf{x}_p\right\Vert_2^2=0\\\frac{1}{C-1}+\frac{1}{\alpha}(a-b)b\left\Vert\mathbf{x}_p\right\Vert_2^2=0\\\left(a^2+(C-1)b^2\right)\left\Vert\mathbf{x}_p\right\Vert_2^2=\alpha.\end{cases}$$Solving these systems of equations we arrive at $$\begin{cases}a=\frac{1}{\left\Vert\mathbf{x}_p\right\Vert_2}\sqrt{\frac{\alpha(C-1)}{C}}\\b=-\frac{1}{\left\Vert\mathbf{x}_p\right\Vert_2}\sqrt{\frac{\alpha}{C(C-1)}}.\end{cases}$$
    \item In the instance of the mean-squared error, $$\ell_p:=\ell\left(f\left(\mathbf{x}_p\right),y_p\right)=\left\Vert f\left(\mathbf{x}_p\right)-\mathbf{e}_{y_p}\right\Vert_2^2,$$where we use $\mathbf{e}_i\in\mathbb{R}^d$ to denote the $i^\text{th}$ standard basis vector. Under the assumptions, the output of the deep network at $\mathbf{x}_p$ is $A_{\omega_{\mathbf{x}_p}}\mathbf{x}_p$. The mean squared error loss of the deep network on $\mathcal{D}$ is $$\mathcal{L}_{\text{MSE}}=\frac{1}{m}\sum_{p=1}^m\ell_p$$where
    \begin{align*}
        \ell_p&=\left\langle A_{\omega_{\mathbf{x}_p}}\mathbf{x}_p-\mathbf{e}_{y_p},A_{\omega_{\mathbf{x}_p}}\mathbf{x}_p-\mathbf{e}_{y_p}\right\rangle=\sum_{c=1}^C\left(\left\langle\left[A_{\omega_{\mathbf{x}_p}}\right]_{c,\cdot},\mathbf{x}_p\right\rangle-\mathbf{1}_{\{y_p=c\}}\right)^2,
    \end{align*}
    which is a convex function on a convex set. Thus we can consider the sufficient Karush-Kuhn-Tucker conditions with Langrange multiplier, $$\mathcal{L}=\mathcal{L}_{\text{MSE}}-\lambda\left(\sum_{c=1}^C\left\Vert\left[A_{\omega_{\mathbf{x}_p}}\right]_{c,\cdot}\right\Vert_2^2-\alpha\right).$$In particular, the Karush-Kuhn-Tucker conditions have the form
    \begin{align}\label{eq:first_kkt_conditions_mse}
        \frac{\partial\mathcal{L}}{\partial\left[A_{\omega_{\mathbf{x}_p}}\right]_{c,\cdot}}&=2\left(\left\langle\left[A_{\omega_{\mathbf{x}_p}}\right]_{c,\cdot},\mathbf{x}_p\right\rangle-\mathbf{1}_{\left\{y_p=c\right\}}\right)\mathbf{x}_p-2\lambda\left[A_{\omega_{\mathbf{x}_p}}\right]_{c,\cdot}.
    \end{align}
    for $c=1,\dots,C$ and 
    \begin{equation}\label{eq:second_kkt_condition_mse}
        \frac{\partial\mathcal{L}}{\partial\lambda}=\sum_{c=1}^C\left\Vert\left[A_{\omega_{\mathbf{x}_p}}\right]_{c,\cdot}\right\Vert_2^2-\alpha=0.
    \end{equation}
    From \eqref{eq:first_kkt_conditions_mse}, we have
    \begin{align*}
        0&=\sum_{c=1}^C\left\langle\left[A_{\omega_{\mathbf{x}_p}}\right]_{c,\cdot},\frac{\partial\mathcal{L}}{\partial\left[A_{\omega_{\mathbf{x}_p}}\right]_{c,\cdot}}\right\rangle\\&=-2\left\langle\left[A_{\omega_{\mathbf{x}_p}}\right]_{y_p,\cdot},\mathbf{x}_p\right\rangle+2\sum_{c=1}^C\left\langle\left[A_{\omega_{\mathbf{x}_p}}\right]_{c,\cdot},\mathbf{x}_p\right\rangle^2-2\lambda\sum_{c=1}^C\left\Vert\left[A_{\omega_{\mathbf{x}_p}}\right]_{c,\cdot}\right\Vert_2^2.
    \end{align*}
    Then, in conjunction with \eqref{eq:second_kkt_condition_mse}, it follows that $$\lambda=-\frac{1}{\alpha}\left\langle\left[A_{\omega_{\mathbf{x}_p}}\right]_{y_p,\cdot},\mathbf{x}_p\right\rangle+\frac{1}{\alpha}\sum_{c=1}^C\left\langle\left[A_{\omega_{\mathbf{x}_p}}\right]_{c,\cdot},\mathbf{x}_p\right\rangle^2.$$Using this back in \eqref{eq:first_kkt_conditions_mse} we obtain,
    \begin{align*}
        \frac{\partial\mathcal{L}}{\partial\left[A_{\omega_{\mathbf{x}_p}}\right]_{c,\cdot}}&=2\left(\left\langle\left[A_{\omega_{\mathbf{x}_p}}\right]_{c,\cdot},\mathbf{x}_p\right\rangle-\mathbf{1}_{\left\{y_p=c\right\}}\right)\mathbf{x}_p\\&+\left(\frac{2}{\alpha}\left\langle\left[A_{\omega_{\mathbf{x}_p}}\right]_{y_p,\cdot},\mathbf{x}_p\right\rangle-\frac{2}{\alpha}\sum_{c=1}^C\left\langle\left[A_{\omega_{\mathbf{x}_p}}\right]_{c,\cdot},\mathbf{x}_p\right\rangle^2\right)\left[A_{\omega_{\mathbf{x}_p}}\right]_{c,\cdot}
    \end{align*}
    To obtain the optimal $A$, it suffices to find $A_{\omega_{\mathbf{x}_p}}$ satisfying these conditions. To do so we consider the ansatz $$\left[A_{\omega_{\mathbf{x}_p}}\right]_{c,\cdot}=\begin{cases}a\mathbf{x}_p&c=y_p\\b\mathbf{x}_p&c\neq y_p.\end{cases}$$Substituting this into our conditions it follows that $b=0$ and $a=\frac{\sqrt{\alpha}}{\left\Vert\mathbf{x}_p\right\Vert_2}$.
\end{enumerate}
\qed

\paragraph{Proof of Theorem \ref{thm:optimally_robust}.}\label{proof:optimally_robust}

Without loss of generality, we can assume $A_{\omega_{\mathbf{x}}}$ to be of the form $\mathbf{c}\mathbf{v}^\top$ for some $\mathbf{v}\in\mathbb{R}^d$. In particular, we assume that the vector $\mathbf{v}$ is of the same norm as $\mathbf{x}$. Then, locally in $\omega_{\mathbf{x}}$, we have $$f_{\theta}(\mathbf{x}+\boldsymbol{\epsilon})=\mathbf{c}\mathbf{v}^\top\left(\mathbf{x}+\boldsymbol{\epsilon}\right).$$Therefore, $\mathbf{x}$ will only be misclassified by the neural network when $\mathbf{v}^\top\left(\mathbf{x}+\boldsymbol{\epsilon}\right) < 0$. From the Cauchy-Schwartz inequality we have that $$-\Vert\mathbf{v}\Vert_2\Vert\boldsymbol{\epsilon}\Vert_2\leq\mathbf{v}^\top\boldsymbol{\epsilon} < -\mathbf{v}^\top\mathbf{x}.$$Hence, $$\Vert\boldsymbol{\epsilon}\Vert_2 > \frac{\mathbf{v}^\top\mathbf{x}}{\Vert\mathbf{v}\Vert_2},$$the right-hand side of which is maximized when $\mathbf{v}$ is $\mathbf{x}$. \qed

\paragraph{Proof of Theorem \ref{thm:pd_parameters}.}\label{proof:pd_parameters}

Using Theorem \ref{thm:centroids_layer} and Theorem \ref{thm:centroids_network}, it follows that 
\begin{align*}
    \mu_{\omega_{\mathbf{x}}^{(1\leftarrow l)}}^{(1\leftarrow l)}&=\left(A^{(l-1)}_{\omega_{\mathbf{x}}^{(l-1)}}\cdots A^{(1)}_{\omega_{\mathbf{x}}^{(1)}}\right)^\top\mu_{\omega_{\mathbf{x}}^{(l)}}^{(l)}\\&=\left(A^{(l-1)}_{\omega_{\mathbf{x}}^{(l-1)}}\cdots A^{(1)}_{\omega_{\mathbf{x}}^{(1)}}\right)^\top\left(A^{(l)}_{\omega_{\mathbf{x}}^{(l)}}\right)^\top\mathbf{1}\\&=\left(A^{(l)}_{\omega_{\mathbf{x}}^{(l)}}\cdots A^{(1)}_{\omega_{\mathbf{x}}^{(1)}}\right)^\top\mathbf{1}\\&=\left(A^{(1\leftarrow l)}_{\omega_{\mathbf{x}}^{(1\leftarrow l)}}\right)^\top\mathbf{1}.
\end{align*}
Extending this to the $L^\text{th}$ yields the desired result. \qed

\paragraph{Proof of Proposition \ref{prop:jacobian_aligned_implies_centroid_aligned}.}\label{proof:jacobian_aligned_implies_centroid_aligned}

Using Theorem \ref{thm:pd_parameters}, the centroid of an aligned Jacobian is $\mu_{\mathbf{x}}=\mathbf{x}\mathbf{c}^\top\mathbf{1}=c\mathbf{x}$ where $c=\mathbf{c}^\top\mathbf{1}$. \qed

\paragraph{Proof of Lemma \ref{lem:centroid_two_layer_network}.}\label{proof:centroid_two_layer_network}

This follows immediately from the application of Theorem \ref{thm:pd_parameters}. \qed

\paragraph{Proof of Lemma \ref{lem:ntk}.}\label{proof:ntk}

Observe that in this setting we have $$\Theta\left(\mathbf{x},\mathbf{x}^\prime\right)=\left\langle\nabla_{W^{(2)}}f_{\theta}(\mathbf{x}),\nabla_{W^{(2)}}f_{\theta}\left(\mathbf{x}^\prime\right)\right\rangle+\left\langle\nabla_{W^{(1)}}f_{\theta}(\mathbf{x}),\nabla_{W^{(1)}}f_{\theta}\left(\mathbf{x}^\prime\right)\right\rangle.$$Therefore, noting that $$\nabla_{W^{(2)}}f_{\theta}(\mathbf{x})=\sigma\left(W^{(1)}\mathbf{x}\right)$$and$$\nabla_{W^{(1)}}f_{\theta}(\mathbf{x})=W^{(2)}Q[\mathbf{x}]\mathbf{x}^\top,$$the result follows. \qed

\paragraph{Proof of Theorem \ref{thm:centroid_alignment_ntk}.}\label{proof:centroid_alignment_ntk}

In a similar way to Proposition \ref{prop:centroid_dynamics}, one can show that     \begin{align*}
    \partial_t\mu_{\mathbf{x}}=&\frac{\eta}{m}\sum_{p=1}^m\Bigg(\left(m\left[\mathbf{x}_p\right]^\top W^{(2)}Q\left[\mathbf{x}_p\right]Q[\mathbf{x}]\left(W^{(2)}\right)^\top\mathbf{1}\right)\mathbf{x}_p\\&\quad\qquad+\left(W^{(1)}\right)^\top Q[\mathbf{x}]\sigma\left(W^{(1)}\mathbf{x}_p\right)m\left[\mathbf{x}_p\right]^\top\mathbf{1}\Bigg).
\end{align*}
Using Lemma \ref{lem:ntk}, this simplifies to $$\partial_t\left(\left\langle\mathbf{x}^\prime,\mu_{\boldsymbol{\nu}}\right\rangle\right)=\frac{\eta}{m}\sum_{p=1}^m\Theta\left(\mathbf{x}^\prime,\mathbf{x}_p\right)m\left[\mathbf{x}_p\right].$$ \qed

\paragraph{Proof of Proposition \ref{prop:centroid_dynamics}.}\label{proof:centroid_dynamics}

From Lemma \ref{lem:centroid_two_layer_network}, observe that $$\partial_t\mu_{\mathbf{x}}=\left(\partial_t\left(W^{(2)}\right)Q[\mathbf{x}]W^{(1)}+W^{(2)}Q[\mathbf{x}]\partial_t\left(W^{(1)}\right)\right)^\top\mathbf{1},$$where $$\partial_t\left(W^{(i)}\right)=-\eta\nabla_{W^{(i)}}\mathcal{L}$$for $i=1,2$. One can show that $$\nabla_{W^{(1)}}\mathcal{L}=-\frac{1}{m}\sum_{p=1}^m\left(W^{(2)}Q\left[\mathbf{x}_p\right]\right)^\top\mathbf{m}\left[\mathbf{x}_p\right]\mathbf{x}_p^\top$$and$$\nabla_{W^{(2)}}\mathcal{L}=-\frac{1}{m}\sum_{p=1}^m\mathbf{m}\left[\mathbf{x}_p\right]\sigma\left(W^{(1)}\mathbf{x}_p\right)^\top.$$Therefore, 
\begin{align*}
    \partial_t\mu_{\mathbf{x}}=&\frac{\eta}{m}\sum_{p=1}^m\Bigg(\left(\mathbf{m}\left[\mathbf{x}_p\right]^\top W^{(2)}Q\left[\mathbf{x}_p\right]Q[\mathbf{x}]\left(W^{(2)}\right)^\top\mathbf{1}\right)\mathbf{x}_p\\&\quad\qquad+\left(W^{(1)}\right)^\top Q[\mathbf{x}]\sigma\left(W^{(1)}\mathbf{x}_p\right)\mathbf{m}\left[\mathbf{x}_p\right]^\top\mathbf{1}\Bigg).
\end{align*}
In particular, $$\mathbf{m}\left[\mathbf{x}_p\right]^\top\mathbf{1}=1-\frac{\sum_{c=1}^C\exp\left(\left[f_{\theta}\left(\mathbf{x}_p\right)\right]_c\right)}{\sum_{c^\prime=1}^C\exp\left(\left[f_{\theta}\left(\mathbf{x}_p\right)\right]_{c^\prime}\right)}=0,$$meaning$$\partial_t\mu_{\mathbf{x}}=\frac{\eta}{m}\sum_{p=1}^m\left(\mathbf{m}\left[\mathbf{x}_p\right]^\top W^{(2)}Q\left[\mathbf{x}_p\right]Q[\mathbf{x}]\left(W^{(2)}\right)^\top\mathbf{1}\right)\mathbf{x}_p.$$Therefore, the result follows since $\partial_t\left\langle\mathbf{x},\mu_{\mathbf{x}}\right\rangle=\left\langle\mathbf{x},\partial_t\left(\mu_{\mathbf{x}}\right)\right\rangle$. \qed

\section{Supporting Results}\label{sec:supporting_results}

Note that, for a continuous piecewise affine network $f=\left(f^{(L)}\circ\dots\circ f^{(1)}\right)$, each $f^{(l)}$ and sub-component $f^{(1\leftarrow l)}=\left(f^{(l)}\circ\dots\circ f^{(1)}\right)$ are also continuous piecewise affine networks. Let $A^{(l)}_{\omega^{(l)}_{\mathbf{x}}}$, $B^{(l)}_{\omega^{(l)}_{\mathbf{x}}}$, $\omega^{(l)}_{\mathbf{x}}$, $\mu^{(\ell)}_{\omega_{\mathbf{x}}^{(l)}}$ and $A^{(1\leftarrow l)}_{\omega^{(1\leftarrow l)}_{\mathbf{x}}}$, $B^{(1\leftarrow l)}_{\omega^{(1\leftarrow l)}_{\mathbf{x}}}$, $\omega^{(1\leftarrow l)}_{\mathbf{x}}$, $\mu_{\omega_{\mathbf{x}}^{(1\leftarrow\ell)}}^{(1\leftarrow l)}$ be analogous notation for the layer and sub-component networks to that of the continuous piecewise affine networks we introduced in Section \ref{sec:jacobian_regularisation_ensures_grokking}.

\begin{theorem}[\citealt{balestriero_geometry_2019}]\label{thm:centroids_layer}
    The $l^\text{th}$ layer of a deep network partitions its input space according to a power diagram with centroids $$\mu^{(\ell)}_{\omega_{\mathbf{x}}^{(l)}}=\left(A^{(l)}_{\omega_{\mathbf{x}^{(l)}}}\right)^\top\mathbf{1}.$$
\end{theorem}

\begin{theorem}[\citealt{balestriero_geometry_2019}]\label{thm:centroids_network}
    The continuous piecewise operation of a deep network from the input to the output of the $l^\text{th}$ layer partitions its input space according to a power diagram with centroids $$\mu_{\omega_{\mathbf{x}}^{(1\leftarrow\ell)}}^{(1\leftarrow l)}=\left(A^{(l-1)}_{\omega_{\mathbf{x}}^{(l-1)}}\cdots A^{(1)}_{\omega_{\mathbf{x}}^{(1)}}\right)^\top\mu_{\omega_{\mathbf{x}}^{(l)}}^{(l)}=:\left(A^{(1\leftarrow l-1)}_{\omega_{\mathbf{x}}^{(1\leftarrow l-1)}}\right)^\top\mu_{\omega_{\mathbf{x}}^{(l)}}^{(l)}.$$
\end{theorem}

\section{Compute Resources}\label{sec:compute_resources}

Our experiments were computed on a range of NVIDIA GPUs, including GTX TITAN Xs, RTX 2080Tis, and RTX 8000s. In Table \ref{tab:compute_resources} we indicate roughly how long each of our main experiments took to run.

\begin{table}[h]
    \centering
    \vspace{1em}
    \caption{\small The computational resources utilised to perform the experiments of this work.}
    \vspace{1em}
    \begin{tabular}{lcr}
        \midrule
        Experiment & Hardware & Time \\
        \midrule
        Figure \ref{fig:pc1} & GTX TITAN X & 4 hours \\ 
        \midrule
        Figure \ref{fig:ntk} & GTX TITAN X & Less than 1 hour \\
        \midrule
        Figure \ref{fig:xor_grokking} & GTX TITAN X & Less than 1 hour \\
        \midrule
        Figures \ref{fig:cnn_delayed_robustness} and \ref{fig:cifar_alignment} & GTX 1080 Ti & 6 hours \\
        \midrule
        Figure \ref{fig:restricted_jacobian} & GTX TITAN X & 4 hours \\
        \midrule
        Figures \ref{fig:transformer_alignment_learnable} and \ref{fig:transformer_alignment_fixed} & GTX 1080 Ti & 5 hours\\
        \midrule
        Figure \ref{fig:scaling} & GTX TITAN X & 1 day \\
        \midrule
        Tables \ref{tab:jr_grokking_improvements} and \ref{tab:adversarial_training} & GTX 1080 Ti, RTX 8000 & 5 days \\
        \midrule
    \end{tabular}
    \label{tab:compute_resources}
\end{table}

\end{document}